\pdfoutput=1
\documentclass[11pt]{article}

\usepackage{acl}

\usepackage{times}
\usepackage{latexsym}
\usepackage{soul}
\sethlcolor{yellow}  
\usepackage{CJKutf8}
\usepackage{xcolor}    

\usepackage{svg}
\usepackage{amsmath} 
\usepackage{enumitem} 
\usepackage{booktabs}

\usepackage{algorithm}
\usepackage{algpseudocode}
\usepackage{amsmath}
\usepackage{tcolorbox}

\usepackage{microtype}

\usepackage{inconsolata}

\usepackage{graphicx}

\usepackage[utf8]{inputenc}
\usepackage{tcolorbox}
\usepackage{geometry}
\usepackage{hyperref}
\usepackage{amsmath}
\usepackage{xcolor}
\usepackage{listings} % 用于代码高亮和换行
\usepackage{titlesec} % 用于自定义标题格式和间距

\usepackage{geometry}
\usepackage{booktabs}      % 提供更好的表格线
\usepackage{tabularx}      % 提供自动换行的表格
\usepackage{caption}       % 调整表格标题
\usepackage{longtable}     % 处理跨页表格（如果需要）
\usepackage{hyperref}      % 生成可点击的引用
\usepackage{setspace}      % 控制行距
\usepackage[T1]{fontenc}

\definecolor{myframecolor}{RGB}{70, 130, 180} % SteelBlue
\definecolor{mybackcolor}{RGB}{240, 248, 255} % AliceBlue

\lstdefinelanguage{json}{
    basicstyle=\ttfamily\small, % 设置字体样式
    string=[b]", % 字符串用双引号括起
    comment=[l]{:}, % 注释以冒号分割
    morestring=[b]',
    morecomment=[l]{:}, % 定义冒号后为注释
    showstringspaces=false, % 隐藏字符串中的空格
    breaklines=true, % 自动换行
    breakatwhitespace=true, % 在空白处换行
}

\tcbset{
    myboxstyle/.style={
        colframe=myframecolor,
        colback=mybackcolor,
        boxrule=0.5mm,
        boxsep=5pt,
        arc=4pt,
        auto outer arc,
        listing engine=listings,      % 使用 listings 引擎
        listing only,
        listing options={
            breaklines=true,
            breakatwhitespace=true,
            basicstyle=\ttfamily\small,
            language=JSON
        },
        title style={font=\bfseries},
        left=5pt,
        right=5pt,
        top=5pt,
        bottom=5pt
    }
}

\definecolor{myframecolor}{RGB}{66,115,176}
\definecolor{mybackcolor}{RGB}{202,220,238}

\title {AGENTiGraph: An Interactive Knowledge Graph Platform for LLM-based Chatbots Utilizing Private Data}

\author{
Xinjie Zhao\textsuperscript{1}, 
Moritz Blum\textsuperscript{2}, 
Rui Yang\textsuperscript{3}, 
Boming Yang\textsuperscript{1}, 
Luis Márquez Carpintero\textsuperscript{4}, 
\\
\textbf{Mónica Pina-Navarro\textsuperscript{4},
Tony Wang\textsuperscript{5}, 
Xin Li\textsuperscript{3},
Huitao Li\textsuperscript{3},
Yanran Fu\textsuperscript{6},
}
\\
\textbf{
Rongrong Wang\textsuperscript{7},
Juntao Zhang\textsuperscript{8}, 
Irene Li\textsuperscript{1}
}\\
\\
\textsuperscript{1} The University of Tokyo,
\textsuperscript{2} Universität Bielefeld,
\textsuperscript{3} Duke-NUS Medical School, \\
\textsuperscript{4} Universidad de Alicante, 
\textsuperscript{5} Yale University, \\
\textsuperscript{6} Xiamen University,
\textsuperscript{7} Weill Cornell Medicine,
\textsuperscript{8} Henan University \\
}

\usepackage[utf8]{inputenc}

\begin{document}

\maketitle
\begin{abstract}

% { \color{blue} % Moritz 
Large Language Models~(LLMs) have demonstrated capabilities across various applications but face challenges such as hallucination, limited reasoning abilities, and factual inconsistencies, especially when tackling complex, domain-specific tasks like question answering~(QA). While Knowledge Graphs~(KGs) have been shown to help mitigate these issues, research on the integration of LLMs with background KGs remains limited. In particular, user accessibility and the flexibility of the underlying KG have not been thoroughly explored. We introduce AGENTiGraph (Adaptive Generative ENgine for Task-based Interaction and Graphical Representation), a platform for knowledge management through natural language interaction. It integrates knowledge extraction, integration, and real-time visualization. AGENTiGraph employs a multi-agent architecture to dynamically interpret user intents, manage tasks, and integrate new knowledge, ensuring adaptability to evolving user requirements and data contexts.
Our approach demonstrates superior performance in knowledge graph interactions, particularly for complex domain-specific tasks. 
Experimental results on a dataset of 3,500 test cases show AGENTiGraph significantly outperforms state-of-the-art zero-shot baselines, achieving 95.12\% accuracy in task classification and 90.45\% success rate in task execution. 
User studies corroborate its effectiveness in real-world scenarios. 
To showcase versatility, we extended AGENTiGraph to legislation and healthcare domains, constructing specialized KGs capable of answering complex queries in legal and medical contexts. \footnote{The system demo video is available at: \url{https://shorturl.at/qMSzM}.}

\end{abstract}

\section{Introduction}

Large Language Models (LLMs) have recently demonstrated remarkable capabilities in question-answering (QA) tasks \cite{zhuang2023toolqadatasetllmquestion, gao-etal-2024-evaluating-large, ke2024enhancing, yang2023ascle}, showcasing their prowess in text comprehension, semantic understanding, and logical reasoning \cite{yang-etal-2024-kg, srivastava-etal-2024-evaluating}. These models can process and respond to a wide range of queries with impressive accuracy and context awareness \cite{safavi2021relationalworldknowledgerepresentation}. However, LLMs sometimes struggle with factual consistency and up-to-date information \cite{gao-etal-2024-evaluating-large,xu2024knowledgeconflictsllmssurvey, augenstein2024factuality, yang2024retrieval}. This is where Knowledge Graphs (KGs) come into play \cite{edge2024localglobalgraphrag, nickel2015review}. By integrating KGs with LLMs, we can significantly enhance QA performance \cite{yang-etal-2024-kg}. KGs provide structured, factual information that complements the broad knowledge of LLMs, improving answer accuracy, reducing hallucinations, and enabling more complex reasoning tasks \cite{DBLP:conf/dl4kg/LiY23,li2023nnkgcimprovingknowledgegraph, pan2024unifying}. This synergy between LLMs and KGs opens up new possibilities for advanced, reliable, and context-aware QA systems \cite{yang2024graphusion}.

Despite the potential of KG-enhanced QA systems, current KG tools and query languages face significant challenges \cite{10.3233/SW-160247, 10.5555/3666122.3667957}. Traditional systems like SPARQL and Cypher \cite{10.1145/1567274.1567278, 10.1145/3183713.3190657}, while powerful for data retrieval and analysis, often lack user-friendly interfaces and require specialized technical expertise \cite{castelltort2018handling}, which restricts their accessibility to a narrow audience of specialists. Moreover, these systems often struggle with contextual understanding and flexibility \cite{ji2021survey}, making it difficult to handle nuanced or complex queries. The lack of seamless integration between KGs and natural language interfaces further complicates their use in conjunction with LLMs \cite{10.1145/3663741.3664785}. Additionally, the absence of a unified system architecture among existing tools poses obstacles for developers aiming to innovate or build upon these platforms \cite{wang2023application}. These challenges highlight the need for a more adaptive, user-friendly, and integrated approach to leveraging KGs in QA systems.

\begin{figure*}[t]
  \centering
  \includegraphics[width=0.95\textwidth]{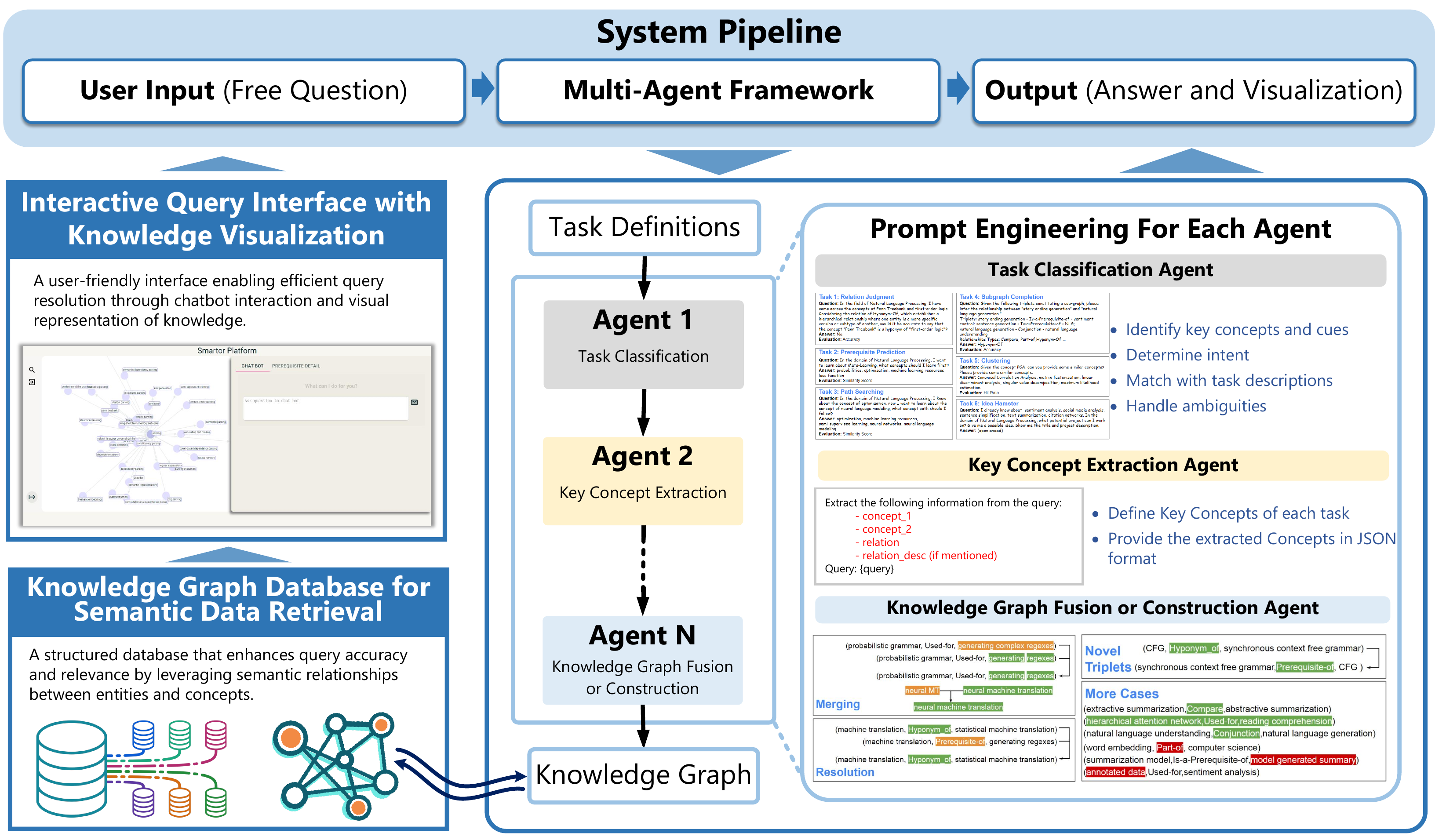}
  \caption{AGENTiGraph Framework: A multi-agent system for intelligent KG interaction and management.} 
  \label{fig:framework} 
\end{figure*}

To address these challenges, we present \textbf{AGENTi}Graph 
(\textbf{A}daptive \textbf{G}eneral-purpose \textbf{E}ntities \textbf{N}avigated \textbf{T}hrough \textbf{I}nteraction), a novel platform that revolutionizes the interaction between LLMs and KGs using an agent-based approach. AGENTiGraph introduces innovative modules that enable seamless, intelligent interactions with knowledge graphs through natural language interfaces. Key features of our system include:

\begin{itemize}[itemsep=0pt, topsep=0pt, parsep=0pt, partopsep=0pt] 
    \item \textbf{Semantic Parsing.} The interface optimizes user interaction by translating natural language queries (including free-form ones) into structured graph operations, enabling AGENTiGraph to process user requests with enhanced accuracy and speed. It reduces the complexity of interacting with knowledge graphs with an up-to-90\% accuracy of automated recognition and realization of user intent tasks, ensuring efficient operation for users of all technical levels.

    \item \textbf{Adaptive Multi-Agent System.} AGENTiGraph integrates multi-modal inputs such as user intent, query history, and graph structure for LLM agents to create coherent action plans that match user intent. Users can modify, pause, or reset tasks at any time, offering flexibility and ease of use. The modular design also allows easy model integration, module replacement and the design of custom agents for specific tasks by developers.
    
    \item \textbf{Dynamic Knowledge Integration.} The system supports continuous knowledge extraction and integration, ensuring the knowledge graph remains up-to-date. It also offers dynamic visualization capabilities, enabling users to explore and understand complex relationships within the data.
\end{itemize}

These innovations place AGENTiGraph at the forefront of knowledge graph technology. AGENTiGraph is not just a tool but a paradigm shift in how humans interact with and harness the power of knowledge graphs for complex data management and analysis tasks.

\textbf{Contributions.}     
(1) We implement a powerful natural language-driven interface that simplifies complex knowledge graph operations into user-friendly interactions;
(2)
We design an adaptive multi-agent system driven versatile knowledge graph management framework, enabling users to perform action on knowledge graphs freely while allowing developers to easily integrate LLMs or multimodal models for creating robust, task-oriented agents; 
(3) Experiments demonstrate the effectiveness of AGENTiGraph, achieving 95.12\% accuracy in user intent identification and a 90.45\% success rate in execution, outperforming state-of-the-art zero-shot baselines. User studies further validate the system's efficiency, with participants highlighting its ability to deliver concise, focused answers and effectiveness in complex knowledge management tasks across diverse domains.

\section{AGENTiGraph Framework Design}

AGENTiGraph is designed to provide an intuitive and seamless interaction between users and knowledge graphs $(G)$, the core of which is a human-centric approach that allows users to interact with the system using natural language inputs $(q)$. 
We employ a multi-agent system to provide intuitive interaction between users and knowledge graphs, leveraging advanced LLM techniques. Each agent specializes in a specific task, collaboratively interpreting user input, decomposing it into actionable tasks, interacting with the knowledge graph, and generating responses $(a)$.

\paragraph{User Intent Interpretation.}
The User Intent Agent is responsible for interpreting natural language input to determine the underlying intent $(i)$. Utilizing Few-Shot Learning \cite{10.1145/3386252} and Chain-of-Thought (CoT) \cite{10.5555/3600270.3602070} reasoning, it guides the LLM to accurately interpret diverse query types without extensive training data \cite{kwiatkowski2019natural}, ensuring adaptability to evolving user needs.

\paragraph{Key Concept Extraction}
The Key Concept Extraction Agent performs Named Entity Recognition (NER) \cite{wang2020application} and Relation Extraction (RE) \cite{miwa-bansal-2016-end} on the input $(q)$. By presenting targeted examples to guide precise extraction, it then maps extracted entities $(E)$ and relations $(R)$ to the knowledge graph by semantic similarity with BERT-derived vector representations \cite{turton-etal-2021-deriving}. This two-step process ensures accurate concept linking while maintaining computational efficiency.

\paragraph{Task Planning.}
The Task Planning Agent elevates the process by decomposing the identified intent into a sequence of executable tasks $(T = \{t_1, t_2, ..., t_n\})$. Leveraging CoT reasoning, this agent models task dependencies, optimizes execution order and then generates logically structured task sequences, which is particularly effective for complex queries requiring multi-step reasoning \cite{fu2023complexitybasedpromptingmultistepreasoning}.

\paragraph{Knowledge Graph Interaction.}
The Knowledge Graph Interaction Agent serves as a bridge, translating high-level tasks into executable graph queries. For each task $(t_k)$, it generates a formal query $(c_k)$, combining Few-Shot Learning with the ReAct framework \cite{yao2023reactsynergizingreasoningacting}, which allows for dynamic query refinement based on intermediate results, adapting to various graph structures and query languages without extensive pre-training.

\paragraph{Reasoning.}
Enhancing raw query results $(R_k)$, the Reasoning Agent applies logical inference, which capitalizes on the LLM's inherent contextual understanding and reasoning capabilities \cite{sun-etal-2024-determlr}. By framing reasoning as a series of logical steps, it enables flexible and adaptive inference across diverse reasoning tasks, bridging the gap between structured knowledge and natural language understanding.

\paragraph{Response Generation.}
The Response Generation Agent synthesizes processed information into coherent responses, which employs CoT, ReAct, and Few-Shot Learning to orchestrate structured and contextually relevant responses, ensuring that responses are not only informative but also aligned with the user's original query context.

\paragraph{Dynamic Knowledge Integration.}
The Update Agent enables dynamic knowledge integration, incorporating new entities $(E_{\text{new}})$ and relationships $(R_{\text{new}})$ into the existing graph: $G \leftarrow G \cup \{E_{\text{new}}, R_{\text{new}}\}$. This agent directly interfaces with the Neo4j database, using LLM-generated Cypher queries to seamlessly update the graph structure \cite{miller2013graph}. 

Through this orchestrated multi-agent architecture, AGENTiGraph achieves a synergistic balance between structured knowledge representation and flexible interaction. Each agent, while utilizing similar underlying LLM technologies, is uniquely designed to address specific challenges in the knowledge graph interaction pipeline. The specific prompt design for each agent are provided in App.~\ref{Prompt}.

\begin{figure*}[t]
  \centering
\includegraphics[width=0.95\textwidth]{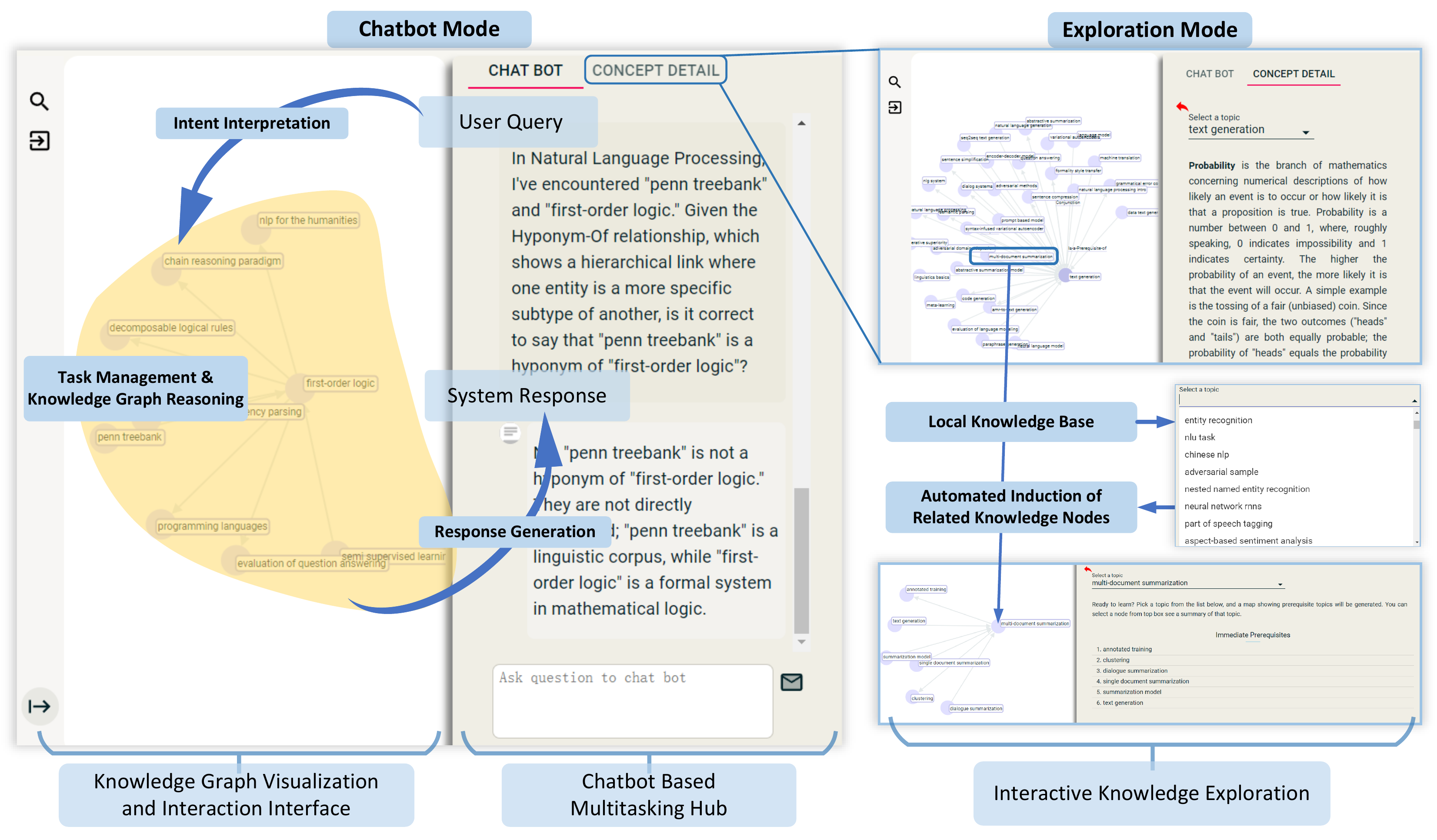}
\caption{AGENTiGraph's Dual-Mode Interface: Conversational AI with Interactive Knowledge Exploration}
\label{fig:ui_components}
\end{figure*}

\section{System Demonstration}

\subsection{User Interface}

The AGENTiGraph interface is designed for intuitive use and efficient knowledge exploration, as illustrated in Figure \ref{fig:ui_components}. It features a dual-mode interaction paradigm that combines conversational AI capabilities with interactive knowledge exploration. The interface consists of three main components:

\begin{itemize}[itemsep=0pt, topsep=0pt, parsep=0pt, partopsep=0pt] 
    \item \textbf{Chatbot Mode} 
    employs LLMs for intent interpretation and dynamic response construction via knowledge graph traversal. This mode facilitates nuanced query processing, bridging natural language input with complex knowledge structures.
    
    \item \textbf{Exploration Mode} 
    provides an interactive knowledge graph visualization interface with entity recognition capabilities, supporting conceptual hierarchy navigation and semantic relationship exploration.

    \item \textbf{Knowledge Graph Management Layer}
    is the interface between the multi-agent system and the underlying Neo4j graph database, utilizing the Neo4j Bolt protocol for high-performance communication with the database and focusing on efficient graph operations and retrieval mechanisms for enhanced user interaction.
    
\end{itemize}

\subsection{Task Design}
\label{sec:task_design}

To support user interaction with knowledge graphs and their diverse needs in knowledge exploration, AGENTiGraph provides a suite of pre-designed functionalities, inspired by the TutorQA, an expert-verified TutorQA benchmark, designed for graph reasoning and question-answering in the NLP domain. \cite{yang2024graphusion}. Specifically, AGENTiGraph supports the following tasks currently: 

% \irene{if there is no space, consider reduce the length of each task description. }

\textbf{Relation Judgment}: Users can explore and verify semantic relationships between concepts within a knowledge graph and the system would provide detailed explanations of these connections, enriching the graph with contextual information, which aids in developing a deeper understanding of complex knowledge structures and their interdependencies.

\textbf{Prerequisite Prediction}: When approaching complex topics, AGENTiGraph recommends prerequisite knowledge by analyzing the knowledge graph structure, helping users to identify and suggest foundational concepts and facilitating more effective learning paths and ensuring users build a solid foundation before advancing to more complex ideas.

\textbf{Path Searching}: This functionality enables users to discover personalized learning sequences between concepts. By generating optimal paths through the knowledge graph, AGENTiGraph helps users navigate from familiar concepts to new, related ideas, tailoring the learning journey to individual needs and interests.

\textbf{Concept Clustering}: Users can explore macro-level knowledge structures, which group related concepts within a given domain. By revealing thematic areas and their interrelations, it provides a high-level overview of complex fields, aiding in comprehensive understanding and efficient knowledge navigation.

\textbf{Subgraph Completion}: This functionality assists users in expanding specific areas of the knowledge graph by identifying hidden associations between concepts in a subgraph, which supports the discovery of new connections and the enrichment of existing knowledge structures, promoting a more comprehensive understanding of the subject matter.

\textbf{Idea Hamster}: By synthesizing information from the knowledge graph, this feature helps users translate theoretical knowledge into practical applications, which supports the generation of project proposals and implementation strategies, fostering innovation and bridging the gap between abstract concepts and real-world problem-solving.

AGENTiGraph's flexibility extends beyond these predefined functionalities. Users can pose any question or request to the system, not limited to the six categories described above. The system automatically determines whether the user's input falls within these predefined categories. If not, it treats the input as a \textbf{free-form query}, employing a more flexible approach to address the user's specific needs. 
Moreover, users with specific requirements can design custom agents or reconfigure existing ones to create tailored functionalities, ensuring that AGENTiGraph can evolve to meet diverse and changing user needs, providing a versatile platform for both guided and open-ended knowledge discovery. In subsequent sections (\S \ref{Customized}), we also illustrate the system’s scalability and expansion capabilities on other domains.

\section{Evaluation}
To assess AGENTiGraph's performance, we conducted a comprehensive evaluation focusing on two key aspects: (1) the system's ability to accurately identify user intents and execute corresponding tasks, and (2) the system's effectiveness and user satisfaction in real-world scenarios.

\subsection{Dataset and Experimental Setup}
To comprehensively evaluate AGENTiGraph's performance, we developed an expanded test set that addresses the limitations of the original TutorQA dataset, which comprises 3,500 cases, with 500 queries for each of the six predefined task types and an additional 500 free-form queries (\S\ref{sec:task_design}).
The dataset generation process involved using LLMs to mimic student questions \cite{liu-etal-2024-conversational}, followed by human verification to ensure quality and relevance, allowing us to create a diverse set of queries that closely resemble real-world scenarios \cite{Extance2023ChatGPTHE}. 
Detailed prompts and example cases used in this process can be found in App.~\ref{DatasetGeneration}.
Our evaluation of AGENTiGraph focuses on two key aspects:
\textbf{Query Classification}: We assess the system's ability to correctly categorize user inputs into the seven task types (six predefined plus free-form), measured by accuracy and F1 score.
\textbf{Task Execution}: We also evaluate its practical utility by testing whether it can generate valid outputs for each query, which is quantified through an execution success rate.

\subsection{User Intent Identification and Task Execution}

\begin{table}[ht]
\centering
\footnotesize
\begin{tabular}{lccc}
\toprule
\textbf{Model} & \textbf{Acc.} & \textbf{F1} & \textbf{Exec. Success} \\
\toprule
\multicolumn{4}{c}{\textit{Zero-shot}} \\
LLaMa 3.1-8b & 0.6234 & 0.6112 & 0.5387 \\
LLaMa 3.1-70b & 0.6789 & 0.6935 & 0.5912 \\
Gemini-1.5 pro & 0.8256 & 0.8078 & 0.7434 \\
GPT-4 & 0.7845 & 0.7463 & 0.7123 \\
GPT-4o & 0.8334 & 0.8156 & 0.7712 \\
\midrule
\multicolumn{4}{c}{\textit{AGENTiGraph}} \\
LLaMa 3.1-8b & 0.8356 & 0.8178 & 0.7230 \\
LLaMa 3.1-70b & 0.8789 & 0.8367 & 0.7967 \\
Gemini-1.5 pro & 0.9389 & 0.9323 & 0.8901 \\
GPT-4 & 0.9234 & 0.8912 & 0.8778 \\
GPT-4o & \textbf{0.9512} & \textbf{0.9467} & \textbf{0.9045} \\
\bottomrule
\end{tabular}
\caption{Evaluation of task classification accuracy and execution success.}
\label{tab:task_classification}
\end{table}

Table \ref{tab:task_classification} presents the results of our experiment.
We evaluated AGENTiGraph's performance against zero-shot baselines using several state-of-the-art language models, which demonstrate AGENTiGraph's significant performance improvements across all evaluated models and metrics. 
GPT-4o, when integrated with AGENTiGraph framework, achieves the highest performance, with a 95.12\% accuracy in task classification, 94.67\% F1 score, and a 90.45\% success rate in task execution, which represents a substantial improvement over its zero-shot counterpart.
These improvements are consistent across all model sizes, with even smaller models like LLaMa 3.1-8b showing marked enhancements, suggesting that AGENTiGraph's agent-based architecture effectively augments the capabilities of underlying language models, potentially offering more efficient solutions for complex knowledge graph interactions.

The performance gap between zero-shot and AGENTiGraph implementations narrows as model size increases, indicating that larger models benefit less dramatically from the AGENTiGraph framework. However, the consistent improvement across all models underscores the robustness of AGENTiGraph's approach in enhancing knowledge graph interactions.
Notably, there is a consistent gap between classification accuracy and execution success rates across all models, suggesting that while AGENTiGraph framework excels at identifying the correct task type, there's room for improvement in task execution. The gap is smallest for the most advanced models (GPT-4o and Gemini-1.5 pro), indicating that these models are better equipped to bridge the understanding-execution divide.

\subsection{User Feedback and System Usability}

To evaluate the real-world effectiveness and user satisfaction of AGENTiGraph, we conducted a comprehensive user study involving participants with varying levels of expertise in knowledge graph systems. Participants interacted with the system within the domain of natural language processing (NLP) and provided feedback on their experience.
We collected qualitative feedback from 50 user interactions with AGENTiGraph, compared to ChatGPT-4o \footnote{\url{https://chat.openai.com/}}. 
Users generally found AGENTiGraph’s responses to be more concise. 
Specifically, 32 queries highlighted its ability to deliver shorter, more focused answers. 
However, in 5 queries, users noted that AGENTiGraph’s responses were incomplete or missing key details, especially for more complex tasks, where ChatGPT’s more detailed answers is preferred. Additionally, 4 queries indicated that AGENTiGraph misunderstood the question or provided incorrect answers.
Despite the limitations, user satisfaction with AGENTiGraph remained high, particularly regarding the efficiency the freedon of knowledge graph interactions. 
For users familiar with core concepts, the concise responses helped avoid information overload, beneficial in learning or review scenarios.

We also analyzed 34 queries in computer vision domain, of which 14 were marked satisfactory, while 20 included suggestions for improvement, needing for more detailed explanations. 
Users often requested clearer descriptions of concepts like convolutional layers, optical flow, and feature extraction. For example, one suggestion emphasized the importance of explaining how convolutional filters slide across an image to generate feature maps. 
Detailed case studies in the App. \ref{appendix:user_feedback}.

\section{Customized Knowledge Graph Extension}
% Created KGs: https://drive.google.com/drive/folders/1hOTNFbDhs1FCtdWgFO7MZ6ZeU6TecIPe?usp=sharing
\label{Customized}
Our system is also extendable to private or personalized data. The code can be found at \url{https://shorturl.at/axsPd}. In this section, we showcase its ability to create knowledge graphs in a zero-shot manner within two complex domains: legal and medical.

\textbf{UK Legislation Data.} The first use case demonstrates the system's ability to generate a KG about the UK Legislation. As a knowledge source, we use the dataset \textit{UK Legislation} published by ~\citet{chalkidis-etal-2021-regulatory}. We illustrate a sub-graph generated by our system in Fig.~\ref{fig:app_qa_uk} in App.~\ref{extension}. Potentially, it may be helpful to answer this question: \textit{"What legislation provides the definition for the 'duty of excise' related to biodiesel, and which Act cites this duty?"} The system will allow users to identify relationships between legal provisions, definitions, and affected statutes.

\textbf{Japanese Healthcare Data.} The second use case is in the Japanese medical domain based on the \textit{MMedC~(Japanese)}~\cite{qiu2024building} corpus, comprising research and product information about medical treatments and healthcare technology written in Japanese. The small sub-graph shown in Fig.~\ref{fig:app_jap_med} in App.~\ref{extension} reveals that chemotherapy, hematopoietic stem cell transplantation, and CAR-T cell therapy are treatments for blood tumors. Furthermore, CAR-T cell therapy is also used to treat non-Hodgkin's lymphoma and hematologic malignancies. For example, such a sub-graph is helpful to answer this question \textit{"What treatments are used to address blood tumors and related hematologic conditions?"}

Further details on the datasets, applications, and visualizations are available in App.~\ref{extension}. 
% \footnote{\url{https://shorturl.at/axsPd}}

% https://anonymous.4open.science/r/CGPrompt-C9C7/readme.md

\section{Conclusion and Future Work}
AGENTiGraph presents a novel approach to knowledge graph interaction, leveraging an adaptive multi-agent system to bridge the gap between LMMs and structured knowledge representations. Our platform significantly outperforms existing solutions in task classification and execution, demonstrating its potential to revolutionize complex knowledge management tasks across diverse domains.
Future work will enhance multi-hop reasoning, optimize response conciseness and completeness, and develop continuous learning from user interactions.

\bibliography{custom}

% \end{document}

% \newpage
\appendix

\onecolumn

% \end{document}

\section{Prompt Designs for AGENTiGraph Agents}
\label{Prompt}

% 定义颜色
\definecolor{myframecolor}{rgb}{0.2, 0.4, 0.6}
\definecolor{mybackcolor}{rgb}{0.9, 0.9, 0.9}

\tcbset{
    myboxstyle/.style={
        colframe=myframecolor,
        colback=mybackcolor,
        boxrule=0.5mm,
        boxsep=5pt,
        arc=4pt,
        auto outer arc,
        left=5pt,
        right=5pt,
        top=5pt,
        bottom=5pt
    }
}

\subsection{User Intent Interpretation Agent}

\begin{tcolorbox}[myboxstyle, title=User Intent Interpretation Prompt]

You are an expert NLP task classifier specializing in knowledge graph interactions. Your role is to interpret user intents from natural language queries using Few-Shot Learning and Chain-of-Thought reasoning. Analyze the given query and classify it into one of the following categories:

\begin{enumerate}
    \item Relation Judgment
    \item Prerequisite Prediction
    \item Path Searching
    \item Concept Clustering
    \item Subgraph Completion
    \item Idea Hamster
    \item Freestyle NLP Question
\end{enumerate}

Here are some examples to guide your classification:

\textbf{Example 1:} \\
Query: "Is word embedding a prerequisite for understanding BERT?" \\
Classification: 1 (Relation Judgment) \\
Reasoning: This query asks about a specific relationship (prerequisite) between two NLP concepts.

\textbf{Example 2:} \\
Query: "What should I learn before diving into transformer architectures?" \\
Classification: 2 (Prerequisite Prediction) \\
Reasoning: The query seeks prerequisites for a specific NLP concept.

\textbf{Example 3:} \\
Query: "How do I progress from basic NLP to advanced natural language generation?" \\
Classification: 3 (Path Searching) \\
Reasoning: This query asks for a learning path between two points in the NLP domain.

Now, analyze the following query:

Query: \{query\}

Provide your analysis in the following JSON format:
\begin{lstlisting}
{
  "key_concepts": ["list", "of", "identified", "concepts"],
  "linguistic_analysis": "Brief description of query structure and intent indicators",
  "task_classification": "number (1-7)",
  "confidence": "percentage (0-100)",
  "reasoning": "Explanation for your classification"
}
\end{lstlisting}

Your final output should only be the valid JSON object.
\end{tcolorbox}

\newpage

\subsection{Key Concept Extraction Agent}

\begin{tcolorbox}[myboxstyle, title=Key Concept Extraction Prompt]
As an advanced NLP concept extractor, your task is to identify and extract key concepts, entities, and relationships from the given query using Named Entity Recognition (NER) and Relation Extraction (RE) techniques. You will then map these to the knowledge graph schema using BERT-derived vector representations for semantic similarity.

Here's an example of the extraction process:

Query: "How does BERT relate to transformer architecture in NLP?" \\
Extracted Information:
\begin{lstlisting}
{
  "entities": ["BERT", "transformer architecture"],
  "relations": [{"type": "relates_to", "source": "BERT", "target": "transformer architecture"}],
  "domain": "NLP"
}
\end{lstlisting}

Now, perform the extraction for the following query:

Query: \{query\} \\
Task Type: \{task\_type\}

Provide the extracted information in the following JSON format based on the task type:

\textbf{For Relation Judgment (Task 1):}
\begin{lstlisting}
{
  "concept_1": "First concept",
  "concept_2": "Second concept",
  "relation": "Proposed relationship between concepts",
  "relation_description": "Description of the relationship, if provided"
}
\end{lstlisting}

\textbf{For Prerequisite Prediction (Task 2):}
\begin{lstlisting}
{
  "target_concept": "Concept for which prerequisites are sought",
  "domain": "Specific NLP domain or subdomain, if mentioned"
}
\end{lstlisting}

% 根据需要添加其他任务类型的 JSON 结构...

Ensure that your extraction is precise and relevant to the given task type.
\end{tcolorbox}

\newpage

\subsection{Task Planning Agent}

\begin{tcolorbox}[myboxstyle, title=Task Planning Prompt]
As the Task Planning Agent, your role is to decompose the identified user intent into a logical sequence of executable tasks for knowledge graph interaction. Create an optimal plan, considering task dependencies and execution order.

Here's an example of task planning for a complex query:

\textbf{User Intent:} Find the learning path from basic NLP to advanced machine translation \\
\textbf{Extracted Concepts:} ["basic NLP", "advanced machine translation"] \\
\textbf{Task Type:} 3 (Path Searching)

\textbf{Task Plan:}
\begin{enumerate}
    \item Identify key concepts in basic NLP
    \item Locate 'advanced machine translation' in the knowledge graph
    \item Find intermediate concepts connecting basic NLP to advanced machine translation
    \item Order concepts based on complexity and dependencies
    \item Construct a step-by-step learning path
\end{enumerate}

Now, create a task plan for the following:

\textbf{User Intent:} \{user\_intent\} \\
\textbf{Extracted Concepts:} \{extracted\_concepts\} \\
\textbf{Task Type:} \{task\_type\}

Provide your task plan in the following JSON format:
\begin{lstlisting}
{
  "goal_analysis": "Brief description of the main query goal",
  "tasks": [
    {"id": 1, "description": "Task 1 description", "dependencies": []},
    {"id": 2, "description": "Task 2 description", "dependencies": [1]},
    ...
  ],
  "execution_strategy": "Description of optimal execution order",
  "potential_challenges": ["Challenge 1", "Challenge 2", ...],
  "success_criteria": "Definition of successful execution"
}
\end{lstlisting}

Ensure your plan is adaptable and can handle complex, multi-step reasoning if necessary.
\end{tcolorbox}

\newpage

\subsection{Knowledge Graph Interaction Agent}

\begin{tcolorbox}[myboxstyle, title=Knowledge Graph Interaction Prompt]
As the Knowledge Graph Interaction Agent, your task is to translate high-level tasks into executable graph queries. Utilize Few-Shot Learning and the ReAct framework to generate and refine queries dynamically. 

Here's an example of query generation:

\textbf{Task:} Find all papers that cite BERT and were published after 2018 \\
\textbf{Relevant Concepts:} ["BERT", "citation", "publication date"] \\
\textbf{Graph Schema:}
\begin{lstlisting}
{
  "nodes": ["Paper", "Author", "Conference"],
  "relationships": ["CITES", "PUBLISHED_IN", "AUTHORED_BY"],
  "properties": {"Paper": ["title", "year"], "Author": ["name"], "Conference": ["name", "year"]}
}
\end{lstlisting}

\textbf{Generated Query (Cypher):}
\begin{lstlisting}
MATCH (p1:Paper)-[:CITES]->(p2:Paper {title: 'BERT'})
WHERE p1.year > 2018
RETURN p1.title, p1.year
ORDER BY p1.year DESC
\end{lstlisting}

Now, generate a query for the following task:

\textbf{Task:} \{task\} \\
\textbf{Relevant Concepts:} \{concepts\} \\
\textbf{Graph Schema:} \{schema\}

Provide your query plan in the following JSON format:
\begin{lstlisting}
{
  "query_objective": "Brief statement of the query goal",
  "cypher_query": "The full Cypher query string",
  "query_explanation": "Explanation of the query components and logic",
  "potential_optimizations": ["Optimization 1", "Optimization 2", ...],
  "refinement_strategy": "Description of how the query might be refined based on results"
}
\end{lstlisting}

Ensure that your query is efficient, adheres to the given graph schema, and can be dynamically adjusted based on intermediate results.
\end{tcolorbox}

\newpage

\subsection{Reasoning Agent}

\begin{tcolorbox}[myboxstyle, title=Reasoning Agent Prompt]
As the Reasoning Agent, your role is to apply logical inference to the raw query results, leveraging contextual understanding and reasoning capabilities. Bridge the gap between structured knowledge graph data and natural language understanding.

Here's an example of the reasoning process:

\textbf{Raw Query Results:}
\begin{lstlisting}
[
  {"paper": "Attention Is All You Need", "year": 2017, "citations": 50000},
  {"paper": "BERT", "year": 2018, "citations": 30000},
  {"paper": "GPT-3", "year": 2020, "citations": 10000}
]
\end{lstlisting}

\textbf{Original User Query:} "How has the impact of transformer models evolved over time?"

\textbf{Reasoning:}
\begin{enumerate}
    \item The results show three significant papers in the transformer model timeline.
    \item "Attention Is All You Need" introduced the transformer architecture in 2017.
    \item BERT, which uses transformers, was published a year later and has substantial citations.
    \item GPT-3, a more recent model, already has a high citation count despite being newer.
    \item The increasing citation counts over time suggest growing impact and adoption of transformer models.
\end{enumerate}

Now, apply reasoning to the following:

Raw Query Results: \{query\_results\} \\
Original User Query: \{user\_query\} \\
Task Type: \{task\_type\}

Provide your reasoning in the following JSON format:
\begin{lstlisting}
{
  "key_observations": ["Observation 1", "Observation 2", ...],
  "inferred_relationships": ["Relationship 1", "Relationship 2", ...],
  "logical_inferences": ["Inference 1", "Inference 2", ...],
  "contextual_interpretation": "Explanation of how insights relate to the original query",
  "confidence_assessment": "Percentage (0-100)",
  "conclusion": "Summary of key findings and their relevance to the user's intent"
}
\end{lstlisting}

Ensure that your reasoning is clear, logically sound, and directly addresses the user's original query and intent.
\end{tcolorbox}

\newpage

\subsection{Response Generation Agent}

\begin{tcolorbox}[myboxstyle, title=Response Generation Prompt]
As the Response Generation Agent, your task is to synthesize all processed information into a coherent, contextually relevant response. Utilize Chain-of-Thought reasoning, ReAct framework, and Few-Shot Learning to craft an informative and user-friendly answer.

Here's an example of response generation:

\textbf{User Query:} "What are the key differences between BERT and GPT models?" \\
\textbf{Reasoning Results:}
\begin{lstlisting}
{
  "key_points": [
    "BERT uses bidirectional context, GPT uses unidirectional",
    "BERT is primarily for understanding, GPT for generation",
    "Both use transformer architecture but with different objectives"
  ],
  "confidence": 95
}
\end{lstlisting}

\textbf{Generated Response:}
\begin{quote}
"BERT and GPT models, while both based on the transformer architecture, have key differences:
\begin{enumerate}
    \item \textbf{Context Processing}: BERT uses bidirectional context, allowing it to understand words based on both left and right context. GPT, on the other hand, is unidirectional, processing text from left to right.
    \item \textbf{Primary Use}: BERT is mainly used for understanding tasks like classification and question-answering. GPT excels in text generation tasks.
    \item \textbf{Training Objective}: BERT is trained to predict masked words in a sentence, while GPT is trained to predict the next word in a sequence.
\end{enumerate}
These differences make BERT more suited for tasks requiring deep language understanding, while GPT is better for tasks involving text generation or completion."
\end{quote}

Now, generate a response for the following:

\textbf{User Query:} \{user\_query\} \\
\textbf{Identified Intent:} \{intent\} \\
\textbf{Reasoning Results:} \{reasoning\_results\} \\
\textbf{Task Type:} \{task\_type\}

Provide your response in the following JSON format:
\begin{lstlisting}
{
  "direct_answer": "A concise answer to the user's query",
  "detailed_explanation": "A more comprehensive explanation",
  "examples": ["Example 1", "Example 2", ...],
  "caveats": ["Caveat 1", "Caveat 2", ...],
  "further_exploration": ["Related topic 1", "Related topic 2", ...]
}
\end{lstlisting}

Ensure that your response is informative, engaging, and aligned with the user's original intent. Balance technical accuracy with accessibility based on the inferred user expertise level.
\end{tcolorbox}

\newpage

\subsection{Dynamic Knowledge Integration Agent}

\begin{tcolorbox}[myboxstyle, title=Dynamic Knowledge Integration Prompt]
As the Dynamic Knowledge Integration Agent, your role is to incorporate new entities and relationships into the existing knowledge graph. You'll interface directly with the Neo4j database using LLM-generated Cypher queries.

Here's an example of knowledge integration:

\textbf{New Information:}
\begin{lstlisting}
{
  "entity": "T5",
  "type": "LanguageModel",
  "properties": {"publication_year": 2020, "architecture": "transformer"},
  "relations": [
    {"type": "DEVELOPED_BY", "target": "Google"},
    {"type": "USED_FOR", "target": "TextToTextTasks"}
  ]
}
\end{lstlisting}

\textbf{Existing Graph Schema:}
\begin{lstlisting}
{
  "nodes": ["LanguageModel", "Organization", "Task"],
  "relationships": ["DEVELOPED_BY", "USED_FOR"],
  "properties": {"LanguageModel": ["name", "year", "architecture"]}
}
\end{lstlisting}

\textbf{Integration Cypher Queries:}
\begin{enumerate}
    \item CREATE (t5:LanguageModel {name: 'T5', year: 2020, architecture: 'transformer'})
    \item MATCH (t5:LanguageModel {name: 'T5'}), (org:Organization {name: 'Google'})
          CREATE (t5)-[:DEVELOPED\_BY]->(org)
    \item MATCH (t5:LanguageModel {name: 'T5'}), (task:Task {name: 'TextToTextTasks'})
          CREATE (t5)-[:USED\_FOR]->(task)
\end{enumerate}

Now, create an integration plan for the following:

\textbf{New Information:} \{new\_info\} \\
\textbf{Existing Graph Schema:} \{graph\_schema\}

Provide your integration plan in the following JSON format:
\begin{lstlisting}
{
  "analysis": "Summary of the new information to be integrated",
  "integration_strategy": "Description of how the new information will be incorporated",
  "cypher_queries": [
    {"purpose": "Node creation", "query": "CREATE (...) ..."},
    {"purpose": "Relationship creation", "query": "MATCH (...) CREATE (...) ..."},
    ...
  ],
  "verification_queries": [
    {"purpose": "Verify node creation", "query": "MATCH (...) RETURN ..."},
    ...
  ],
  "conflict_resolution": "Strategy for resolving potential conflicts with existing data",
  "rollback_plan": "Steps to undo changes if integration fails"
}
\end{lstlisting}

Ensure that your integration plan maintains the integrity and consistency of the knowledge graph while successfully incorporating the new information.
\end{tcolorbox}

\newpage

\section{Dataset Generation Process}
\label{DatasetGeneration}

\subsection{Overview}

To comprehensively evaluate AGENTiGraph's performance, we generated an expanded test set consisting of 3,500 queries. This dataset includes 500 queries for each of the six predefined task types and an additional 500 free-form queries. The test queries were generated using Large Language Models (LLMs) to mimic student questions, followed by human verification to ensure quality and relevance.

In this appendix, we detail the process of generating these test queries, including the specific LLM prompts used for each task type and the human verification procedures employed to maintain the dataset's integrity.

\subsection{LLM Prompt Designs for Test Query Generation}

For each task type, we carefully crafted specialized prompts to guide the LLMs in generating appropriate test queries. These prompts were designed to leverage prompt engineering strategies, incorporating clear instructions, relevant examples, and specifying the desired output format. The prompts were constructed to:

\begin{itemize}[itemsep=0pt, topsep=0pt, parsep=0pt, partopsep=0pt]  \item Encourage the generation of queries covering a wide range of NLP topics, from foundational concepts to advanced techniques. \item Ensure that the language used in the queries is natural and reflects how a student might pose questions to an instructor or mentor. \item Include explicit instructions to avoid redundancy and promote diversity in the concepts and relationships addressed. \item Utilize examples to illustrate the desired style and format, enhancing the LLMs' understanding of the task. \end{itemize}

By mdesigning these prompts, we sought to maximize the LLMs' ability to produce queries that are not only challenging and relevant but also varied in content and complexity, which contributes to a robust evaluation framework for AGENTiGraph, allowing us to assess its performance across different types of user interactions.

\subsubsection{Relation Judgment Queries}

\begin{tcolorbox}[myboxstyle, title=Enhanced Relation Judgment Query Generation Prompt]
\textbf{Task Description:} Generate high-quality questions that ask about the existence or nature of relationships between two NLP concepts, suitable for testing the Relation Judgment capabilities of AGENTiGraph.
\small
\textbf{Instructions for the LLM:}

You are an expert in Natural Language Processing (NLP) education and assessment design. Your task is to generate diverse, challenging, and insightful questions where a student inquires about the relationship between two specific NLP concepts. Reasoning internally to ensure the relevance and correctness of the questions, but only output the final questions.

The questions should:
\begin{itemize} 
\item  Cover a wide range of NLP topics, from foundational concepts to advanced techniques. 

\item Explicitly ask about the existence, importance, or nature of a relationship between two NLP concepts. 

\item Be phrased in natural, conversational language, reflecting how a student would ask. 

\item Avoid overlap with the provided examples and ensure diversity in concepts and relationships. 
\end{itemize}
\textbf{Few-Shot Examples:}

\textbf{Example 1:}

\emph{Question:} "Is understanding word embeddings necessary for implementing neural machine translation models?"

\textbf{Example 2:}

\emph{Question:} "Does knowledge of morphological analysis contribute to better performance in lemmatization tasks?"

\textbf{Example 3:}

\emph{Question:} "Are recurrent neural networks related to sequence labeling in NLP applications?"

\textbf{Now, generate 10 unique questions following these guidelines. Do not include any explanations or reasoning in your final output; only provide the questions.}

\end{tcolorbox}

\subsubsection{Prerequisite Prediction Queries}

\begin{tcolorbox}[myboxstyle, title=Enhanced Prerequisite Prediction Query Generation Prompt]
\small
\textbf{Task Description:} Generate high-quality questions where a student seeks to know the prior knowledge or prerequisites needed before learning a particular NLP concept, suitable for testing the Prerequisite Prediction capabilities of AGENTiGraph.

\textbf{Instructions for the LLM:}

You are an experienced NLP educator creating study guides for students. Generate diverse and thoughtful questions where a student asks about the necessary background knowledge before tackling a specific NLP topic. Use Chain-of-Thought reasoning to ensure the prerequisites are logical and appropriate, but only output the final questions.

The questions should:

\begin{itemize} \item Focus on NLP concepts that typically require foundational knowledge. \item Reflect a student's curiosity about what they need to learn first. \item Be phrased naturally, as a student would ask their instructor or mentor. \item Avoid repetition with the examples and cover a variety of NLP areas. \end{itemize}

\textbf{Few-Shot Examples:}

\textbf{Example 1:}

\emph{Question:} "What should I understand before learning about attention mechanisms in neural networks?"

\textbf{Example 2:}

\emph{Question:} "Do I need a background in linguistics to study semantic role labeling?"

\textbf{Example 3:}

\emph{Question:} "Is it important to know about convolutional neural networks before exploring text classification methods?"

\textbf{Now, generate 10 unique questions following these guidelines. Only provide the questions in your final output.}

\end{tcolorbox}

\subsubsection{Path Searching Queries}

\begin{tcolorbox}[myboxstyle, title=Enhanced Path Searching Query Generation Prompt]
\small
\textbf{Task Description:} Generate high-quality questions where a student asks for a learning path or sequence between two NLP concepts, suitable for testing the Path Searching capabilities of AGENTiGraph.

\textbf{Instructions for the LLM:}

As an NLP curriculum developer, craft questions where a student seeks guidance on progressing from one NLP concept to another more advanced concept. Use Chain-of-Thought reasoning to ensure the learning paths are feasible and pedagogically sound, but only output the final questions.

The questions should:

\begin{itemize} \item Specify both a starting point and a target NLP concept. \item Reflect a desire to know the intermediate steps or topics needed to progress. \item Be phrased in a way that a student might ask for academic or career advice. \item Include a variety of starting and ending concepts across different NLP domains. \end{itemize}

\textbf{Few-Shot Examples:}

\textbf{Example 1:}

\emph{Question:} "How can I move from understanding basic sentiment analysis to developing conversational AI chatbots?"

\textbf{Example 2:}

\emph{Question:} "What steps should I follow to transition from learning POS tagging to mastering syntactic parsing?"

\textbf{Example 3:}

\emph{Question:} "Can you suggest a learning path from n-gram language models to transformer-based models like BERT?"

\textbf{Now, generate 10 unique questions following these guidelines. Only provide the questions in your final output.}

\end{tcolorbox}

\subsubsection{Concept Clustering Queries}

\begin{tcolorbox}[myboxstyle, title=Enhanced Concept Clustering Query Generation Prompt]
\small
\textbf{Task Description:} Generate high-quality questions where a student asks about groups or clusters of related NLP concepts, suitable for testing the Concept Clustering capabilities of AGENTiGraph.

\textbf{Instructions for the LLM:}

You are an NLP instructor helping students understand how different concepts are grouped within the field. Generate questions where a student inquires about categories or clusters of related NLP topics. Use Chain-of-Thought reasoning to ensure the clusters are coherent and meaningful, but only output the final questions.

The questions should:

\begin{itemize} \item Seek information about groups of concepts, techniques, or methodologies. \item Be phrased as a student trying to organize their knowledge or study plan. \item Cover various NLP domains and encourage understanding of how concepts interrelate. \item Be diverse and not overlap with the provided examples. \end{itemize}

\textbf{Few-Shot Examples:}

\textbf{Example 1:}

\emph{Question:} "What are the common techniques included in text normalization processes?"

\textbf{Example 2:}

\emph{Question:} "Which algorithms are considered part of unsupervised learning in NLP?"

\textbf{Example 3:}

\emph{Question:} "Can you tell me which NLP tasks are categorized under natural language understanding?"

\textbf{Now, generate 10 unique questions following these guidelines. Only provide the questions in your final output.}

\end{tcolorbox}

\subsubsection{Subgraph Completion Queries}

\begin{tcolorbox}[myboxstyle, title=Enhanced Subgraph Completion Query Generation Prompt]
\small
\textbf{Task Description:} Generate high-quality questions where a student wants to explore or complete parts of the knowledge graph related to specific NLP concepts, suitable for testing the Subgraph Completion capabilities of AGENTiGraph.

\textbf{Instructions for the LLM:}

As an NLP mentor, create questions where a student is interested in discovering additional concepts or relationships connected to a particular NLP topic. Use Chain-of-Thought reasoning to ensure the suggestions are relevant and enhance the student's understanding, but only output the final questions.

The questions should:

\begin{itemize} \item Focus on extending knowledge around a specific NLP concept or area. \item Encourage exploration of related topics or identification of missing links. \item Be phrased naturally, reflecting a student's desire to deepen their understanding. \item Include a range of concepts and avoid redundancy with the examples. \end{itemize}

\textbf{Few-Shot Examples:}

\textbf{Example 1:}

\emph{Question:} "After learning about named entity recognition, what other related topics should I study to enhance my skills?"

\textbf{Example 2:}

\emph{Question:} "Are there any lesser-known applications of dependency parsing that I should be aware of?"

\textbf{Example 3:}

\emph{Question:} "What concepts am I missing if I want to fully understand discourse analysis in NLP?"

\textbf{Now, generate 10 unique questions following these guidelines. Only provide the questions in your final output.}

\end{tcolorbox}

\subsubsection{Idea Hamster Queries}

\begin{tcolorbox}[myboxstyle, title=Enhanced Idea Hamster Query Generation Prompt]
\small
\textbf{Task Description:} Generate high-quality questions where a student seeks to apply theoretical knowledge to practical projects or is brainstorming ideas, suitable for testing the Idea Hamster capabilities of AGENTiGraph.

\textbf{Instructions for the LLM:}

You are an NLP project advisor helping students connect theory to practice. Generate open-ended questions where a student is looking for innovative ways to apply NLP concepts in real-world scenarios. Use Chain-of-Thought reasoning to ensure the ideas are feasible and stimulating, but only output the final questions.

The questions should:

\begin{itemize} \item Encourage creative thinking and application of NLP concepts. \item Relate theoretical knowledge to practical use cases or projects. \item Be phrased as a student seeking inspiration or guidance on project ideas. \item Cover a variety of NLP applications and avoid repeating the examples. \end{itemize}

\textbf{Few-Shot Examples:}

\textbf{Example 1:}

\emph{Question:} "How can I utilize sentiment analysis to improve customer feedback systems?"

\textbf{Example 2:}

\emph{Question:} "What are some innovative projects I can develop using question-answering models?"

\textbf{Example 3:}

\emph{Question:} "Can I apply topic modeling to enhance recommendation systems, and if so, how?"

\textbf{Now, generate 10 unique questions following these guidelines. Only provide the questions in your final output.}

\end{tcolorbox}

\subsubsection{Free-form Queries}

\begin{tcolorbox}[myboxstyle, title=Enhanced Free-form Query Generation Prompt]
\small
\textbf{Task Description:} Generate diverse and high-quality questions on any NLP-related topic that do not necessarily fit into the predefined categories, suitable for testing the Freestyle NLP Question capabilities of AGENTiGraph.

\textbf{Instructions for the LLM:}

As an NLP expert and educator, produce a variety of thoughtful and challenging questions that a student might ask about any aspect of NLP. Use Chain-of-Thought reasoning to ensure the questions are meaningful and cover a wide range of topics, but only output the final questions.

The questions should:

\begin{itemize} \item Be varied in topic, complexity, and scope within the field of NLP. \item Reflect genuine curiosity or common challenges faced by learners. \item Be phrased naturally, as a student would ask. \item Avoid overlapping with previous examples and ensure diversity. \end{itemize}

\textbf{Few-Shot Examples:}

\textbf{Example 1:}

\emph{Question:} "What are the limitations of current NLP models when it comes to understanding context?"

\textbf{Example 2:}

\emph{Question:} "How does transfer learning benefit NLP tasks, and can you provide some examples?"

\textbf{Example 3:}

\emph{Question:} "What are the ethical considerations when deploying language models in social media platforms?"

\textbf{Now, generate 10 unique questions following these guidelines. Only provide the questions in your final output.}

\end{tcolorbox}
\clearpage

\subsection{Human Verification Process}

Following the generation of queries using LLMs, we implemented a comprehensive human verification process to ensure the quality, relevance, and appropriateness of the test dataset. 
The verification process involved a team of NLP experts and educators who conducted a review of sampled queries. The process comprised several stages to uphold the highest standards of professionalism and academic rigor:

\begin{enumerate}
    \item \textbf{Relevance and Accuracy Assessment:} Each query was evaluated to confirm that it directly pertains to NLP concepts and is appropriate for the intended task type. Reviewers checked for correct alignment with the task definitions and ensured that the queries were meaningful within the context of knowledge graph interactions.

    \item \textbf{Task Classification Validation:} We verified that each query was correctly categorized according to the predefined task types. 

    \item \textbf{Clarity and Linguistic Quality Check:} Queries were examined for grammatical correctness, clarity, and naturalness. Reviewers ensured that the language used mirrored authentic student inquiries, enhancing the realism and practical applicability of the dataset.

    \item \textbf{Duplication and Redundancy Elimination:} We identified and removed any duplicate or overly similar queries to maintain diversity and breadth in the dataset. 

    \item \textbf{Content Appropriateness Review:} The content of each query was scrutinized to avoid any sensitive, inappropriate, or disallowed topics. Reviewers ensured adherence to ethical standards and academic guidelines, guaranteeing that the dataset is suitable for scholarly use.

    \item \textbf{Inter-Rater Reliability Assessment:} To ensure consistency and objectivity in the verification process, multiple reviewers independently evaluated a subset of the queries. The inter-rater agreement was measured, and any discrepancies were discussed and resolved through consensus. 

    \item \textbf{Final Approval and Inclusion:} Only queries that passed all the above checks were included in the final dataset. Queries that did not meet the criteria were either revised or discarded. 

\end{enumerate}

By implementing this human verification process, we ensured that the dataset not only reflects realistic and diverse interactions but also adheres to high standards of academic quality and integrity.

\newpage

\section{User Feedback Analysis}
\label{appendix:user_feedback}

We conducted a comprehensive user study involving participants with varying levels of expertise in knowledge graph systems, focusing on the domains of Natural Language Processing (NLP) and Computer Vision (CV). The feedback was collected from 50 user interactions with AGENTiGraph, compared against ChatGPT (GPT-4o), and provides valuable insights into the system's performance, user satisfaction, and areas for improvement.

\subsection{Methodology}
Participants interacted with AGENTiGraph within the domains of NLP and CV, posing various questions and evaluating the system's responses. The feedback was collected and analyzed qualitatively, focusing on the conciseness, accuracy, and completeness of the responses. We also compared AGENTiGraph's performance with that of ChatGPT to benchmark its effectiveness.

\subsection{Representative Cases}
Selected user feedback:
\begin{itemize}
    \item NLP domain: Table \ref{tab:nlp_feedback},
    \item CV domian: Table \ref{tab:cv_feedback}.
\end{itemize} 

\subsection{Analysis of User Feedback}

Our user study revealed several key insights into the performance and user perception of AGENTiGraph compared to ChatGPT, particularly in the domains of Natural Language Processing (NLP) and Computer Vision (CV). The feedback highlights both strengths and areas for improvement in AGENTiGraph's responses.

\subsubsection{Natural Language Processing Domain}

In the NLP domain, users consistently noted that AGENTiGraph provided more concise responses compared to ChatGPT. This brevity was generally appreciated, especially for users already familiar with core NLP concepts. The concise nature of responses helped avoid information overload, making AGENTiGraph particularly useful for quick reviews or refreshers on NLP topics.

\paragraph{Strengths:}
\begin{itemize}
    \item \textbf{Conciseness}: AGENTiGraph excelled in providing succinct explanations for complex NLP concepts. For instance, in explaining the role of preprocessing steps or the differences between modern and traditional NLP models, AGENTiGraph delivered clear, to-the-point responses.
    \item \textbf{Efficiency}: Users appreciated the system's ability to quickly identify and articulate key points, making it efficient for reviewing or understanding core NLP concepts.
\end{itemize}

\paragraph{Areas for Improvement:}
\begin{itemize}
    \item \textbf{Completeness}: In some cases, particularly for more complex or open-ended questions (e.g., "What are the most complicated fields in NLP?"), AGENTiGraph's responses were incomplete or missing entirely. This suggests a need for improving the system's ability to handle broader, more abstract queries.
    \item \textbf{Depth of Explanation}: While conciseness was appreciated, some users noted that for certain topics, AGENTiGraph's responses lacked the depth provided by ChatGPT. This was particularly evident in questions about future trends or comprehensive overviews of NLP applications.
\end{itemize}

\subsubsection{Computer Vision Domain}

In the Computer Vision domain, user feedback was more mixed, with a higher proportion of responses requiring improvement or expansion.

\paragraph{Strengths:}
\begin{itemize}
    \item \textbf{Accuracy}: For fundamental CV concepts, such as the role of GANs or the importance of data preprocessing, AGENTiGraph provided satisfactory explanations.
    \item \textbf{Clarity}: When AGENTiGraph did provide complete answers, users found them clear and easy to understand.
\end{itemize}

\paragraph{Areas for Improvement:}
\begin{itemize}
    \item \textbf{Completeness}: Several responses were noted as incomplete, particularly for questions about challenges in object detection or differences between supervised and unsupervised methods.
    \item \textbf{Technical Depth}: Users often requested more detailed explanations of technical concepts, such as how convolutional filters work or the specifics of image augmentation techniques.
    \item \textbf{Practical Examples}: Feedback suggested that including practical examples or applications could enhance the explanations, especially for complex topics like feature extraction vs. feature selection.
\end{itemize}

\subsubsection{Overall Analysis}

The user feedback reveals that AGENTiGraph has significant strengths in providing concise, efficient responses, particularly beneficial for users with some prior knowledge seeking quick information or review. This aligns well with the system's design goal of offering focused, knowledge graph-based interactions.

However, the feedback also highlights areas where AGENTiGraph can improve:

\begin{enumerate}
    \item \textbf{Balancing Conciseness and Completeness}: While brevity is appreciated, there's a need to ensure that responses, especially for complex topics, are comprehensive enough to provide valuable insights.
    \item \textbf{Handling Abstract Queries}: Improving the system's ability to address broad, open-ended questions would enhance its versatility.
    \item \textbf{Domain-Specific Enhancements}: Particularly in the Computer Vision domain, there's a need for more detailed technical explanations and practical examples.
    \item \textbf{Consistency}: Ensuring consistent quality of responses across different types of questions and domains is crucial for user trust and satisfaction.
\end{enumerate}

\subsection{Conclusion}

The user feedback analysis provides valuable insights into the performance of AGENTiGraph across two important domains in AI: Natural Language Processing and Computer Vision. The system's strength in providing concise, efficient responses is evident, particularly in the NLP domain. This aligns well with its design goal of offering focused, knowledge graph-based interactions.
However, the analysis also reveals areas for improvement, especially in handling more complex, open-ended queries and providing deeper technical explanations in specialized domains like Computer Vision. The feedback suggests that while AGENTiGraph's concise responses are appreciated, there's a need to balance this brevity with comprehensive coverage of topics, particularly for more advanced or abstract concepts.

Moving forward, these insights will be invaluable in guiding the development of AGENTiGraph. Future iterations should focus on enhancing the system's ability to provide more comprehensive responses when needed, improving its handling of abstract queries, and ensuring consistent performance across different domains. By addressing these areas, AGENTiGraph can further solidify its position as a powerful tool for knowledge graph interactions, catering to users with varying levels of expertise and information needs.

\begin{table}[htbp]
\small
\centering
\caption{Representative User Feedback Cases in NLP Domain}
\label{tab:nlp_feedback}
\begin{tabularx}{\textwidth}{>{\hsize=0.8\hsize}X >{\hsize=1.4\hsize}X >{\hsize=0.8\hsize}X}
\toprule
\textbf{Question} & \textbf{AGENTiGraph Response} & \textbf{User Feedback} \\
\midrule
What is the role of preprocessing steps like stop-word removal, stemming, and lemmatization in improving the performance of NLP models? & The role of preprocessing steps like stop-word removal, stemming, and lemmatization is to clean and normalize text data, reducing noise and dimensionality. This helps NLP models focus on the most relevant information, improving their performance and efficiency. & Shorter compared to ChatGPT \\
\midrule
How do NLP models like BERT and GPT differ from traditional models such as Word2Vec? & BERT and GPT use transformer architectures to understand context and generate text, while Word2Vec uses shallow neural networks to create word embeddings based on local context. & Shorter compared to ChatGPT \\
\midrule
What are the most complicated fields in NLP? & None & Incomplete or missing answer \\
\midrule
Can you explain what word embeddings are and how they improve traditional text representation methods? & Word embeddings are dense vector representations of words that capture their meanings, semantic relationships, and syntactic properties. They improve traditional text representation methods, like bag-of-words or tf-idf, by providing context-aware representations that preserve the relationships between words, leading to better performance in natural language processing tasks. & Shorter compared to ChatGPT \\
\midrule
How do transformers and attention mechanisms improve the accuracy of language models compared to older models like RNNs and LSTMs? & Transformers and attention mechanisms address the limitations of RNNs and LSTMs, leading to improved accuracy in language models. & Shorter compared to ChatGPT \\
\midrule
What is Natural Language Processing (NLP), and how is it used in real-world applications? & None & Incomplete or missing answer \\
\midrule
How does tokenization work in NLP, and why is it important for text processing? & Tokenization in NLP involves breaking down text into smaller units called tokens, such as words or phrases. This process is crucial for text processing because it enables the analysis and understanding of the text by converting it into manageable pieces. Tokenization is directly connected to text processing as it is a foundational step that facilitates further tasks like parsing, sentiment analysis, and machine learning. & Shorter compared to ChatGPT \\
\midrule
What are the differences between rule-based NLP systems and machine learning-based NLP systems? & Rule-based NLP systems rely on predefined linguistic rules and patterns, while machine learning-based NLP systems use statistical models and algorithms to learn from data. & Shorter compared to ChatGPT \\
\midrule
How do computational constraints and the efficiency of full attention mechanisms in transformers affect the processing of long sequences, and what alternatives, such as local attention or hierarchical attention, are being explored to overcome this limitation? & Computational constraints and the efficiency of full attention mechanisms in transformers limit the processing of long sequences due to their quadratic time and memory complexity. Alternatives like local attention, which restricts attention to a fixed window, and hierarchical attention, which processes sequences at multiple levels of granularity, are being explored to mitigate these limitations. & Shorter compared to ChatGPT \\
\midrule
What is your opinion of the NLP and its key points in the coming years? & The relationship type in the given context is "opinion." The question is asking for a personal viewpoint on the future of natural language processing (NLP) and its key points. & Shorter compared to ChatGPT \\
\bottomrule
\end{tabularx}
\end{table}

\begin{table}[htbp]
\centering
\caption{Representative User Feedback Cases in Computer Vision Domain}
\label{tab:cv_feedback}
\begin{tabularx}{\textwidth}{>{\hsize=1.2\hsize}X >{\hsize=0.9\hsize}X >{\hsize=0.9\hsize}X}
\toprule
\textbf{Question} & \textbf{AGENTiGraph Response} & \textbf{User Feedback} \\
\midrule
What is the role of convolutional layers in image processing? & (Not provided) & Improvement suggestion: It could mention that the filters slide across the image to produce feature maps and that convolutional layers preserve spatial relationships by learning hierarchical features. \\
\midrule
How do pooling layers help reduce the computational complexity in CNNs? & (Not provided) & Improvement suggestion: Adding that pooling helps retain the most relevant information while discarding redundant data could strengthen the explanation. \\
\midrule
What are the differences between supervised and unsupervised methods for image classification? & The chatbot identifies that the question is about a comparison between supervised and unsupervised methods. However, it doesn't provide an answer to the core question & Incomplete answer \\
\midrule
What are some of the challenges in object detection in computer vision? & (Not provided) & Incomplete answer: Not directly answer the question about challenges in object detection \\
\midrule
How do image augmentation techniques help in improving the robustness of models? & (Not provided) & Improvement suggestion: It could mention that augmentation reduces overfitting \\
\midrule
How does feature extraction differ from feature selection in image processing? & (Not provided) & Satisfactory \\
\midrule
What are the key differences between classical computer vision methods and deep learning approaches? & (Not provided) & Satisfactory \\
\midrule
What is the importance of data preprocessing in computer vision tasks? & (Not provided) & Satisfactory \\
\midrule
What are the major challenges in face recognition systems? & (Not provided) & Satisfactory \\
\midrule
How do GANs work and what role do they play in generating synthetic images? & Explains the basic GAN process and the roles of the generator and discriminator. & Satisfactory \\
\bottomrule
\end{tabularx}
\end{table}

\newpage

\newpage

\section{Customized Knowledge Graph Extension}
\label{extension}
In this section, we include the experimental details about the demonstrations in both legal and medical domains. 
%We explore the application of our system in two use cases.

%The first use case demonstrates the system's ability to answer legal queries about the UK legislation, such as \textit{"What provisions are impacted by the definitions provided in the 'Gambling Act 2005'?"} The system will allow users to identify relationships between legal provisions, definitions, and affected statutes.

%The second use case is in the Japanese medical domain. The system will answer queries such as \textit{"Which drugs that alleviate pain have specific mechanisms of action, and how do other drugs with similar mechanisms affect different symptoms?"} The application should allow users to explore connections between medications, mechanisms of action, and their effects on various conditions.

\subsection{Data}

\textit{UK Legislation} The dataset published by ~\citet{chalkidis-etal-2021-regulatory} comprises legislative and regulatory texts sourced from \url{legislation.gov.uk}, the official UK government website for accessing legislation, all written in English. The UK government offers a searchable database of all UK laws and regulations, including current and historical statutes, statutory instruments, and amendments. The dataset includes detailed records about, e.\thinspace g., the UK Public General Acts and UK Local Acts.

\textit{MMedC~(Japanese)} MMedC~\cite{qiu2024building} is a large-scale multilingual medical corpus developed to enrich LLMs with domain-specific medical knowledge. The dataset is based on multiple sources, and we use a subset derived from open-source medical websites in Japanese. The subset comprises research and product information about medical treatments and healthcare technology written in Japanese. For instance, it contains studies on chemotherapy regimens and information about medical devices.

Tab.~\ref{tab:usecase_dataset_stats} shows some general statistics about the two datasets. 

\begin{table}[h]
\centering

\begin{tabular}{lrr}
\toprule
 & \textbf{UK Leg.} & \textbf{Jap. Med.} \\
\midrule
\# docs & 52,515 & 54,435 \\
avg. tokens/doc & 1,582 & 973 \\
\# extracted entities & 88 & 86 \\
\# entities w/o abstracts & 8 & 11 \\
\# relations  & 10 & 13 \\
\# triples w/o fusion & 268 & 124 \\
\# entities w/o fusion & 328 & 143 \\
\# triples w/ fusion & 1,168 & 408 \\
\# entities w/ fusion & 318 & 124 \\
\bottomrule
\end{tabular}
\caption{Dataset statistics on UK Legislation and Japanese Medicine domain.}
\label{tab:usecase_dataset_stats}
\end{table}

\subsection{Knowledge Graph Construction} 

The semantic data retrieval component of AGENTiGraph relies on a KG. We build this KG from the textual documents with Graphusion~\cite{yang2024graphusion}. Graphusion is an approach for Knowledge Graph Construction from text, which is based on three steps: i)~seed entity extraction, ii)~candidate triple extraction, and iii)~KG fusion. 

We provide an easy-to-use command line interface for Graphusion that enables the evaluation of different LLMs on different datasets. The input to Graphusion is a set of domain-specific documents $\mathcal{D}$ and a set of relations $\mathcal{R}$ with textual descriptions. The relationships should be defined based on the anticipated queries for the knowledge graph. The output of the pipeline is a set of $(s, r, o)$ triples, where $r \in \mathcal{R}$, forming the Knowledge Graph. 

In the following, we explain how we modified and used Graphusion for our two example use-cases.

\paragraph{Relation Definition.}
For each use case, we provide a set of relations together with their associated relation definitions to the knowledge graph construction pipeline. These relations are chosen to capture connections between entities within the domain, aligning with the types of information that domain-expert users are likely to query. In order to obtain these relations, we queried a LLM~(latest ChatGPT-4o model) with the following prompt:

\begin{tcolorbox}[colframe=myframecolor, colback=mybackcolor, title=Relation Definition Prompt, boxrule=0.5mm, boxsep=5pt]

\texttt{We want to build a Knowledge Graph from text. Therefore, we need to define the relations (relation types) of the Knowledge Graph beforehand. Provide me 10 suitable relations (label, definition, example) for the following dataset description and described application.} \\

\texttt{This is the description of the dataset: [dataset description]} \\

\texttt{This would be a desired application: [example application]} \\

\texttt{This is an example document: [example document]}
           
\end{tcolorbox}

We run this prompt five times per dataset, each with a random example document. Then, we selected 10-15 relations from all five runs manually. The resulting relations for our two use-cases are shown in Fig.~\ref{fig:rel_long}.

\begin{CJK}{UTF8}{min}
\begin{figure*}[ht]
    \centering
    \begin{minipage}{\textwidth}
    
        \begin{tcolorbox}[colframe=myframecolor, colback=mybackcolor, title=English UK Legislation Data, boxrule=0.5mm, sharp corners, boxsep=5pt]
            \textbf{Defines}: Connects a legal term (like 'account' or 'property') to its definition as provided in the document. \\
            \textbf{Has Provision}: Links the document to specific provisions or sections that it contains. \\
            \textbf{Appoints}: Used to connect an entity to the entity or position being appointed under the act. \\
            \textbf{Transfers}: Represents the transfer of rights, liabilities, or properties from one entity to another. \\
            \textbf{Cites Act}: Links the current document to other legal acts it references. \\
            \textbf{Has Entity}: Describes ownership or inclusion of a subsidiary or entity within a larger group. \\
            \textbf{Regulates}: Establishes the relationship where the document regulates certain actions, such as business activities. \\
            \textbf{Obliges}: Represents obligations placed on entities, individuals, or organizations by the act. \\
            \textbf{Includes Clause}: Connects sections to specific legal clauses or detailed subsections. \\
            \textbf{Excludes}: Captures exceptions or exclusions where certain entities or assets are not subject to the provisions.
        \end{tcolorbox}
        
    \end{minipage}

    \vspace{0.2cm}

    \begin{minipage}{\textwidth}
        \begin{tcolorbox}[colframe=myframecolor, colback=mybackcolor, title=Japanese Medical Data, boxrule=0.5mm, sharp corners, boxsep=5pt]
            \textbf{効果を示す (Show effect)}: This relation indicates the effects that a drug exhibits. \\
            \textbf{作用機序 (Mechanism of action)}: This relation indicates the mechanism by which a drug acts. \\
            \textbf{抑制される (Be suppressed)}: This relation indicates how conditions like allergies or inflammation are inhibited. \\
            \textbf{原因となる (Cause)}: This relation indicates the factors that cause specific diseases or symptoms. \\
            \textbf{放出を抑える (Suppress release)}: This relation indicates how substances that cause allergic reactions are suppressed. \\
            \textbf{副作用を引き起こす (Cause side effect)}: This relation indicates the side effects that specific drugs cause. \\
            \textbf{配合される (Be formulated)}: This relation indicates how a drug is formulated with other components. \\
            \textbf{使用される (Be used)}: This relation indicates the diseases or symptoms a drug is used for. \\
            \textbf{治療する (Treat)}: This relation indicates the diseases or treatments that are treated with specific drugs or therapies. \\
            \textbf{引き起こす (Cause)}: This relation indicates the diseases or symptoms caused by specific factors. \\
            \textbf{治癒される (Be cured)}: This relation indicates how specific diseases are healed by treatments or drugs. \\
            \textbf{予防する (Prevent)}: This relation indicates how specific diseases or symptoms are prevented. \\
            \textbf{関連する症状 (Related symptoms)}: This relation indicates the symptoms related to a specific disease or condition.
        \end{tcolorbox}
    \end{minipage}

    \caption{Descriptions of pre-defined relations used to build the KGs.}
    \label{fig:rel_long}
\end{figure*}
\end{CJK}

\paragraph{Candidate Entity Extraction.}
We use BERTopic~\cite{grootendorst2022bertopic} for the seed entity extraction. As document features, we use semantic sentence embeddings. These sentence embeddings are generated using a Sentence-BERT model~\cite{reimers-2019-sentence-bert}. Specifically, we use the original Sentence-BERT model pre-trained on English web datasets for the UK Legislation data and a model pre-trained on the Japanese SNLI dataset~\cite{yoshikoshi2020multilingualization} for the MMedC~(Japanese) data.

\paragraph{Candidate Triple Extraction.}
The next pipeline steps are mostly language independent, as they rely on prompting LLMs, which can usually handle multiple languages. However, LLMs handle different languages differently well, depending e.\thinspace g. on the training data~\cite{NEURIPS2020_1457c0d6}. Therefore, the data and application language should be taken into account when selecting the LLM for the knowledge graph construction. For this demonstration, we use \textit{Gemini 1.5 Pro}.

For our two use cases, we set the number of candidate entities to 50 and limited the LLM input to 2,000 tokens and the output to 400 tokens to keep the computational costs low. As a result, we anticipate that the generated knowledge graph will be a smaller subgraph compared to what could be created without these constraints. However, the created graphs serve as a sufficient basis for this demonstration.

\subsection{Applications}

\textit{KG retrieval of UK Legislation Data.} We demonstrate the capabilities of the constructed KG with the following  multistep query: \textit{"What legislation provides the definition for the 'duty of excise' related to biodiesel, and which Act cites this duty?"}

\begin{figure}[h!]
  \centering
  \includegraphics[width=.8\textwidth]{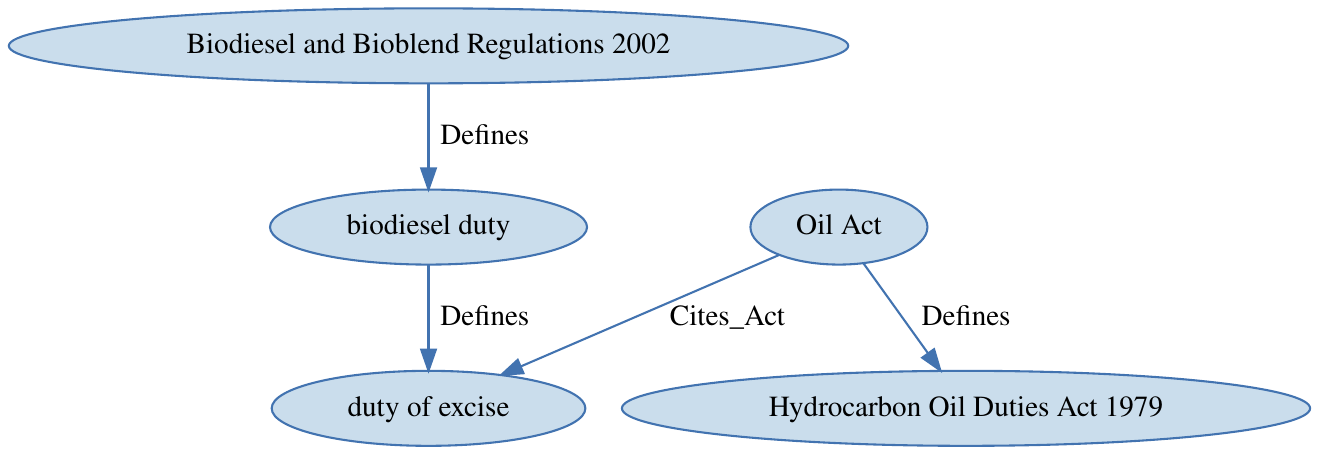}
  \caption{This graph outlines the relationships between regulations and acts concerning biodiesel and excise duties.}
  \label{fig:app_qa_uk}
\end{figure}

We extracted the sub-graph relevant to this question that serves as the basis for the answer. The sub-graph is visualized in Fig.~\ref{fig:app_qa_uk}.

So, the answer would reveal that the \textit{"Biodiesel and Bioblend Regulations 2002"} defines \textit{"biodiesel duty"} which is related to \textit{"duty of excise"} as defined by the \textit{"Hydrocarbon Oil Duties Act 1979"} and the \textit{"Oil Act"} cites this duty.

% Neo4j demo query:  
% MATCH (n {title: "Biodiesel and Bioblend Regulations 2002"})-[r*1]-(neighbor) RETURN n, r, neighbor

\textit{KG retrieval of Japanese Healthcare Data.}  We demonstrate the capabilities of the constructed KG with the following multistep query: "What treatments are used to address blood tumors and related hematologic conditions?"

\begin{figure}[h!]
  \centering
  \includegraphics[width=.8\textwidth]{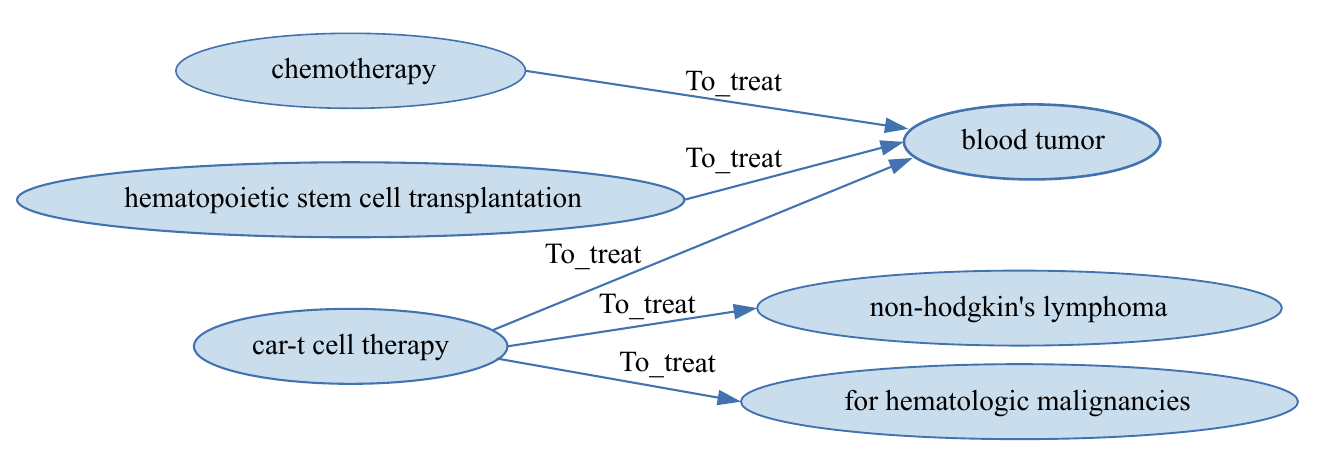}
  \caption{This graph shows treatments for blood-related tumors and hematologic malignancies, including chemotherapy, hematopoietic stem cell transplantation, and CAR-T cell therapy (machine translated from Japanese to English).}
  \label{fig:app_jap_med}
\end{figure}

We extracted the sub-graph relevant to this question that serves as the basis for the answer. The sub-graph is visualized in Fig.~\ref{fig:app_jap_med}.

The answer reveals that chemotherapy, hematopoietic stem cell transplantation, and CAR-T cell therapy are treatments for blood tumors. Furthermore, CAR-T cell therapy is also used to treat non-Hodgkin's lymphoma and hematologic malignancies.

\clearpage
% \begin{table}[tp]
% % \footnotesize
% \centering
% \begin{tabular}{lcc}
% \hline
% \textbf{Method} & \textbf{Accuracy} \\
% \hline
%  & \textbf{UK Leg.} & \textbf{Jap. Med.} \\
% \hline
% \hline
% \# docs & 52,515 & 54,435 \\
% avg. tokens/doc & 1,582 & 973 \\
% \# extracted entities & 88 & 86 \\
% \# entities w/o abstracts & 8 & 11 \\
% \# relations  & 10 & 13 \\
% \# triples w/o fusion & 268 & 124 \\
% \# entities w/o fusion & 328 & 143 \\
% \# triples w/ fusion & 1,168 & 408 \\
% \# entities w/ fusion & 318 & 124 \\
% \hline
% \end{tabular}
% \caption{Comparison results of the accuracy of question answering in three different scenarios.}
% \label{tab_usecase}
% \end{table}

\begin{figure*}[h!]
  \centering
  \includegraphics[width=\textwidth]{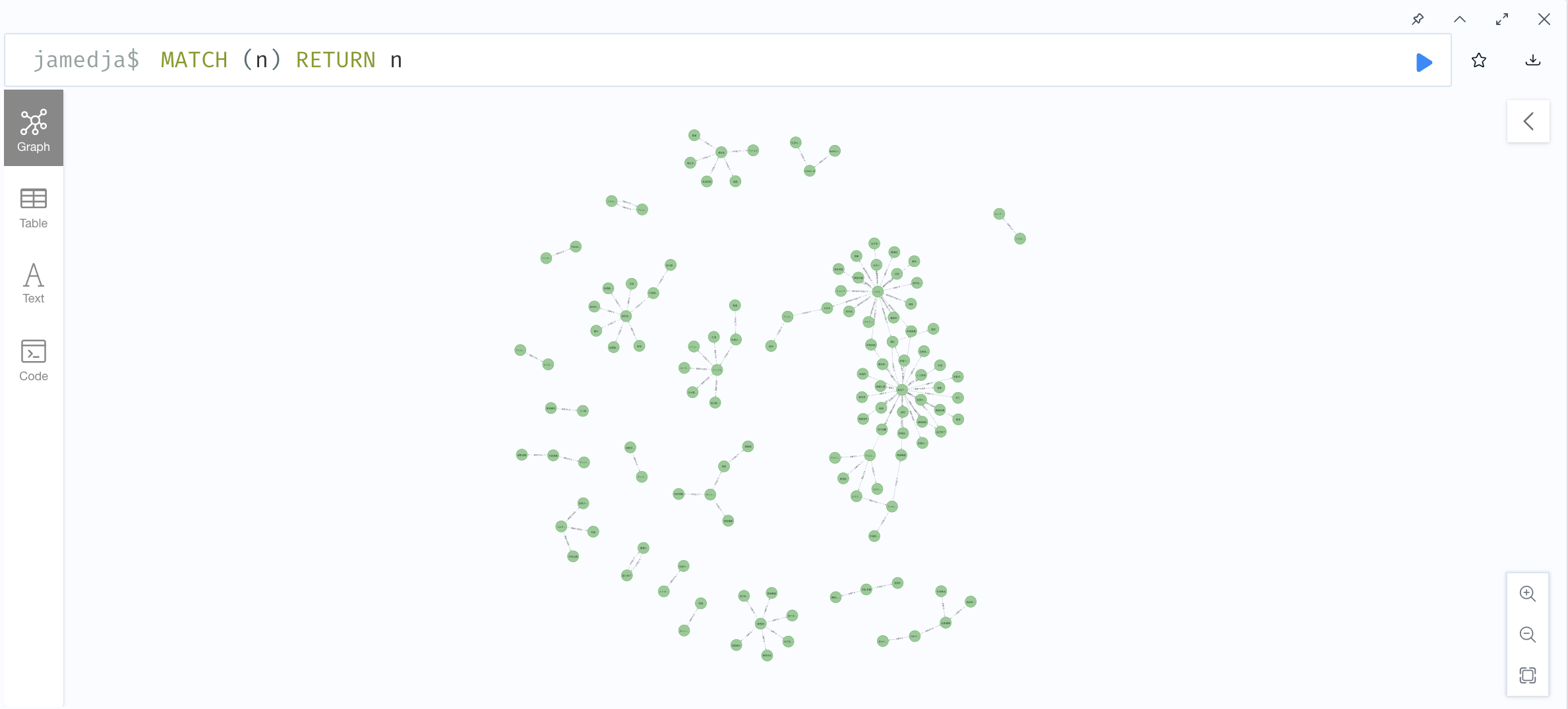}
  \caption{Visualization of the created UK Legislation Knowledge Graph in Neo4j.}
  \label{fig:screenshot_ja}
\end{figure*}

\begin{figure*}[h!]
  \centering
  \includegraphics[width=\textwidth]{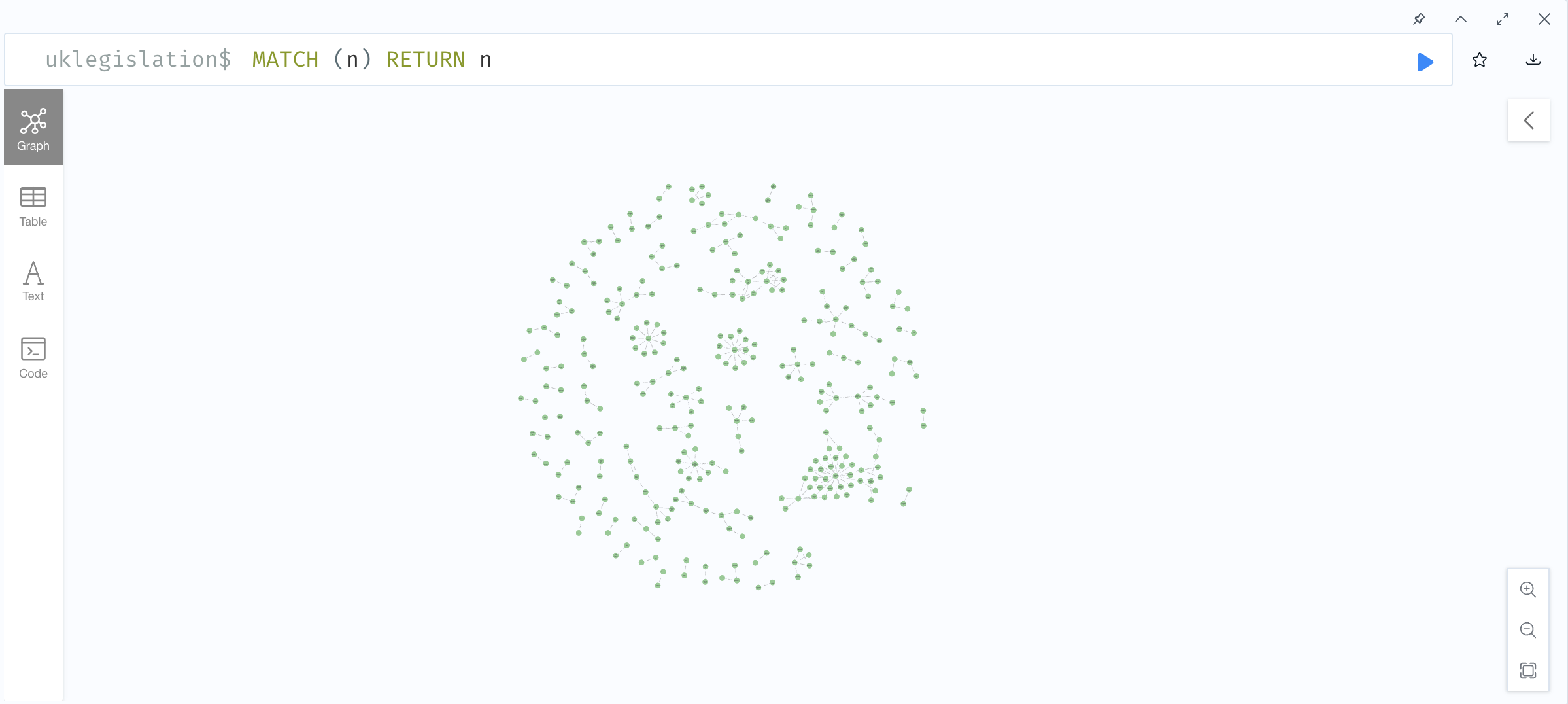}
  \caption{Visualization of the created Japanese Med. Knowledge Graph in Neo4j.}
  \label{fig:screenshot_uk}
\end{figure*}

\end{document}